\documentclass[preprint,12pt,authoryear]{elsarticle}
\usepackage{xcolor}
\usepackage{booktabs}
\usepackage{url}
\usepackage{float}
\def\tsc#1{\csdef{#1}{\textsc{\lowercase{#1}}\xspace}}
\tsc{WGM}
\tsc{QE}
\tsc{EP}
\tsc{PMS}
\tsc{BEC}
\tsc{DE}

\begin{document}
\let\WriteBookmarks\relax
\def\floatpagepagefraction{1}
\def\textpagefraction{.001}




\title{Avatar-Based Conversational AI Assistants in Engineering Education: Towards Process Automation in Chemical Engineering}




%

\author[1]{Fiammetta Caccavale\corref{cor1}}
\cortext[cor1]{Corresponding author}


\ead{fiacac@kt.dtu.dk}



\affiliation[1]{organization={Process and Systems Engineering Center (PROSYS), Department of Chemical and Biochemical Engineering, Technical University of Denmark},
    city={Kongens Lyngby},
    postcode={DK-2800}, 
    country={Denmark}}

\affiliation[2]{organization={Danish Polymer Centre (DPC), Department of Chemical and Biochemical Engineering, Technical University of Denmark},
    city={Kongens Lyngby},
    postcode={DK-2800}, 
    country={Denmark}}

\affiliation[3]{organization={Pilot Plant, Department of Chemical and Biochemical Engineering, Technical University of Denmark},
    city={Kongens Lyngby},
    postcode={DK-2800}, 
    country={Denmark}}

\author[1]
{Carina L. Gargalo}


\author[2]{Magdalena Skowyra}


\author[3]{Julian Kager}


\author[3]{Steen Larsen}


\author[1]
{Krist V. Gernaey}


\author[1]{Ulrich Kr{\"u}hne}



\begin{abstract}
The rapid adoption of conversational artificial intelligence (AI), i.e., technologies such as chatbots and virtual agents that allow users to interact with computers using natural language, in higher education is reshaping established teaching models and student support systems. Although enthusiasm for these tools has spread between teachers and students, critical gaps remain regarding the measurable impact of these tools on student retention and academic performance. 

This study investigates the pedagogical efficacy of replacing traditional human instructors with AI-driven avatars in a specialized audit role-play exercise. We evaluate student performance and perceptions across three distinct environments: a human-led, text-based chatbot, and an avatar-based interface. Our findings demonstrate that all three modalities achieve statistically comparable academic outcomes. Despite this parity in high-level success, qualitative analysis reveals marked divergences in user experience; while students identified higher levels of "authentic immersion" and effective, although complex, information-gathering in human-led settings, they expressed a stronger preference for the future integration of AI and avatar-based tools despite concerns regarding algorithmic hallucinations and structural execution. These results indicate that while AI-driven tools successfully match traditional benchmarks, their implementation requires a fundamental redesign of instructional scaffolding, including enhanced prompt engineering training and \textcolor{black}{carefully defined learning objectives}. Beyond the classroom, this research highlights the broader applicability of conversational AI for professional training, onboarding, and automation within the process and pharmaceutical industries. This work provides a scalable framework for the responsible, human-centric deployment of AI technologies in both educational and industrial contexts.

\end{abstract}


\maketitle

\section{Introduction}
\label{intro}
{\color{black}
In recent years, conversational artificial intelligence (AI) has shifted from a new and experimental classroom tool to a widespread part of higher education's infrastructure. Commercially available tools such as ChatGPT, Copilot and Claude are starting to be widely used \citep{long2026artificial}. These tools have been used increasingly as intelligent tutoring systems, in adaptive learning environments, for automated assessment, and as student-support tools \citep{okonkwo2021chatbots}. Research has shown that these tools can lead to an enhanced learning experience and overall higher engagement of the students \citep{kasneci2023chatgpt}. 

However, there are also many challenges introduced by this technology that educators are facing, such as over-reliance on LLMs \citep{abd2023large}, which could hinder critical thinking and reasoning skills \citep{xu2024opportunities}. \cite{wollny2021we} argue that the educational value of a chatbot depends on more than its ability to perform a task or answer a query; its performance must align with the instructional purpose, the subject matter, and the learners' needs. Recent systematic reviews reinforce that the educational efficacy of conversational AI is fundamentally mediated by instructional design rather than model fluency alone; without deliberate scaffolding, unaligned chatbot interactions fail to sustain higher-order cognitive engagement \citep{long2026artificial}. Empirical studies show that greater fluency in conversational AI can enhance the quality of short-term outputs while also fostering "metacognitive laziness", in which learners skip crucial self-regulation processes when targeted instructional constraints are not in place \citep{fan2025beware}.
Researchers also identify persistent challenges in evaluating educational chatbots and determining how they can support learning beyond basic information exchange \citep{wollny2021we, long2026artificial}.


\textcolor{black}{Thus, while it has been shown that conversational AI can have a positive impact on higher education, rigorous and continuous research is necessary to ensure that its introduction benefits learning and students in general. For example,  the use of AI in the university course connected to this study led to a raise of the enrollment limit in the course, allowing a 66.6\% increase in student uptake, without compromising the teaching quality or reducing the activities planned for the course \citep{caccavale2026large}. In this case, the introduction of AI appeared to have a positive impact on higher education. At the same time, continuing to improve the initiative and further evaluating long-term effects of conversational AI on learning and how it impacts students, especially across cohorts and student populations, is essential to effectively assess a new technology}. 
Therefore, this work aims to continue improving the initiative and further assess the effects of conversational AI on learning. 

This study examines conversational AI in a M.Sc. Good Manufacturing Practice (GMP) course taught at the Technical University of Denmark (DTU). The course includes a learning activity in which students conduct a simulated audit exercise.} 
The primary objective of this study is to introduce a conversational AI model that interacts with students through an avatar. \textcolor{black}{The tool is} created in both a virtual reality (VR) environment and in a 2D Desktop version. This research expands on previous studies performed by the authors, in which a generative AI chatbot, ChatGMP \citep{caccavale2025chatgmp, caccavale2026large}, using retrieval-augmented generations (RAG), was created to simulate an interview-like role-play exercise between students (asking questions) and a fictional company (originally represented by course teachers).
The role of the students in this exercise is to gather information and documents issued by the company. \textcolor{black}{Students' learning is assessed through a written group report.} 

In the study, we compare students' post-activity perceptions across three distinct auditing environments: human-led (Teacher), chatbot (ChatGMP) and the new avatar-based conversational AI tool (Avatar). 
{\color{black}
The avatar-based conversational AI assistant developed in this study is a conversational agent presented through a virtual character that enables students to interact with the system using spoken natural language. 
Comparing these settings offers an opportunity to examine how students perceive AI's general capabilities and how they evaluate AI-mediated interaction in a specific, task-oriented educational context. The anonymous survey administered examines students' general attitudes toward AI, their views on AI use in consequential decisions, their assessment of AI's ability to perform a GMP-like audit, and their perceptions of AI performance relative to a human. It also documents students' general AI-use frequency and their use of AI when preparing audit questions. The study therefore addresses the following research questions:
\begin{enumerate}
    \item What do students think about AI in general?
    \item How comfortable are students with AI making critical decisions without human oversight?
    \item To what extent do students think AI can perform a task like the GMP course audit?
    \item How do students perceive AI performance on these tasks relative to human performance?
    \item How frequently do students use AI?
    \item Do students use AI to prepare their audit questions, and how do they incorporate it into this process?
\end{enumerate}

By addressing these questions, the study contributes context-specific evidence on student perceptions of conversational AI in an applied educational activity. The findings are intended to inform the design and evaluation of AI-supported learning activities, rather than to establish that one auditee format causes better learning outcomes.
}


While the results of the study are collected in an educational environment and the tools are specifically created for this particular scope, we also discuss the relevance of conversational AI technologies to the wider process industry, presenting different tasks this technology can be applied to, such as for employee training and onboarding, as well as for automation of reports, for example in the pharmaceutical sector. 

\section{Background}
\label{background}
\textcolor{black}{This section aims to briefly introduce the current background on conversational AI models embedded in higher education. First, the opportunities that these models could bring to the field will be discussed, followed by an overview of the unique advantages introduced by avatar-based and VR tools.}

\subsection{Conversational AI in education}
Conversational AI, have significantly transformed higher education by making learning more personalized, adaptive, and scalable \citep{labadze2023role}. Powered by large language models (LLMs), these systems address a wide range of pedagogical requirements, including delivering immediate feedback, supporting language learning, and streamlining administrative workflows and content creation \citep{stoehr2024a, rodrigues2024a}. 

\textcolor{black}{These advancements have created new opportunities in higher education. Reported applications include personalized learning \citep{Gan2023Large}, tailored student support \citep{rodrigues2024assessing}, greater teaching efficiency \citep{Xu2024Large}, increased student engagement \citep{Pelaez-Sanchez2024The}, research support, and improved accessibility and inclusivity \citep{Gan2023Large,Xu2024Large}. These applications may contribute to more engaging learning experiences \citep{kasneci2023chatgpt}, while research has also associated AI-supported learning with improved comprehension and academic outcomes \citep{rodrigues2024assessing}. AI-based approaches may further support learning efficiency by helping manage cognitive load, adapting instruction to individual needs, and adjusting learning pathways \citep{gkintoni2025challenging}. Such support may be particularly relevant in cognitively demanding fields such as STEM, where it has been associated with improvements in understanding, retention, and problem-solving \citep{gkintoni2025challenging}. Taken together, this research points to a growing role for LLMs as learning aids in engineering education \citep{kamalov2023new}.}

Yet, their growing use also raises substantial concerns, such as threats to academic integrity, the emergence of algorithmic bias, and the possibility that over-reliance on AI could undermine students' critical thinking and reasoning abilities \citep{abd2023large, xu2024opportunities}. 
\textcolor{black}{According to \citep{kasneci2023chatgpt}, generative AI can introduce risks that are consequential in educational settings. Language models can produce inaccurate or misleading information, reflect biases in their training data, or present uncertain responses with undue confidence. Their use raises further concerns about privacy, transparency, academic integrity, and student reliance on AI-generated content \citep{kasneci2023chatgpt}. These concerns are especially relevant in activities where students must assess information rather than simply retrieve it. Accordingly, UNESCO states that AI-supported learning activities should be designed to encourage verification, critical judgment, and appropriate human oversight  \citep{unesco}.}

Consequently, researchers argue that such technologies should function as collaborative partners that augment the work of human educators rather than replace them, calling for a reorientation toward human-centered design within educational systems. \cite{yan2024practical} provide recommendations for studies involving conversational AI tools, including updating existing innovations with state-of-the-art models, embracing the initiative of open-sourcing models/systems, and adopting a human-centered approach throughout the developmental process.

\subsection{VR in education and its broader application to chemical engineering}
{\color{black}

Virtual reality (VR) has recently being actively incorporated into education, teaching, and training in various application domains \citep{radianti2020systematic}.
\cite{radianti2020systematic} argue that although VR is not a new technology, the recent developments in immersive technologies, in terms of visualization and interactions, have made VR increasingly attractive to scholars. This is also due to the fact that VR head-mounted displays (HMDs), allow users to experience a high degree of immersion (i.e. the involvement of a user in a virtual environment during which their awareness of time and the real world often becomes disconnected, thus providing a sense of "being" in the task environment) \citep{radianti2020systematic}. Due to the increasing research interest in VR technologies, systematic mappings of VR applications for education have emerged, both focusing on desktop VR, such as in \cite{merchant2014effectiveness} and immersive VR \citep{jensen2018review, radianti2020systematic}.

Multi-sensory environments, such as virtual reality (VR) and avatar-based conversational AI tools, can offer unique opportunities for teaching and learning by providing immersive, and context-rich experiences that are difficult to replicate in traditional classroom settings \citep{ding2024integration}. 
Various literature supports the pedagogical effects of augmented or virtual reality technologies; for example, \cite{quintero2019augmented} investigates educational inclusion, highlighting that this technology can have a positive impact on motivation, interaction and generating interest on the part of the student. \cite{di2020review} found that virtual classroom environments as supplementary to traditional teaching environments offers substantial benefits to cognitive and skill‐based learning outcomes. VR or avatars in metaverse-like environments can enable contextual instruction, role-playing, simulations, and collaborative problem-solving, thereby enhancing learner engagement and educational outcomes \citep{lampropoulos2024virtual}. \cite{kumar2021virtual} discusses the pedagogical implications of virtual reality for chemical engineering education.
In this field, this technology can replace costly physical laboratories and intricate real-world environments, expanding access and lowering expenses while still providing practical, hands-on experience \citep{edwards2019haptic}. 
}

Outside traditional educational settings, AI is reshaping industrial operations through smarter process control, predictive maintenance and digitalization, while simultaneously changing job roles and required skills \citep{babashahi2024ai}. AI can be used to customize onboarding programs, tailoring content by role and using chatbots/virtual assistants for rapid Q\&A, while augmenting human interaction \citep{brown2024impact}. Broader workforce reviews emphasize comprehensive, inclusive AI upskilling and lifelong professional development as essential for readiness in AI‑intensive industries \citep{joshi2025agentic}.

\subsubsection{Avatar-based conversational AI Assistant}
{\color{black}
In this work, although both the immersive VR and the desktop VR version of the tool are available, we will mainly focus on the latter, since this was the version students actively tried in the course. To avoid confusion for the readers, we use the terminology of avatar-based conversational AI assistant, instead of refering to the tool as a form of desktop VR.
We define an avatar-based conversational AI as an \textit{AI conversational agent presented through a visual, computer-generated character that interacts with users using natural language, often through spoken dialogue}. Unlike a text-only chatbot, it combines conversational interaction with an on-screen avatar. 

Due to the more recent developments in conversational AI, research focusing specifically on the intersection of avatars(or VR) and AI is sparse. 
Some examples can be found in \cite{so2023natural}, \cite{rashik2024beyond} and \cite{kober2026ai}.
However, we believe that when combined with AI, these virtual (or multi-sensory) environments can be used to further tailor the learning experience, facilitating more flexible, autonomous, and personalized forms of interaction. Thus, further context-specific evidence is needed on how students perceive avatar-based conversational AI assistants, how these tools relate to human interaction.

}

\section{Methods}
\label{methods}

\textcolor{black}{These opportunities and challenges motivate the evaluation of conversational AI in educational context. In this work, we evaluate conversational AI in a specific educational contexts, rather than assuming that its general capabilities translate directly into effective learning support. 
This study therefore compares post-activity perceptions and reported AI use among students who completed the audit with a teacher, a \textcolor{black}{chatbot}, or an avatar. It investigates students' views of AI and their experience with the activity; it does not directly measure learning gains or establish the objective accuracy of the systems, but it presents the performance of students in the final exam to show that the auditee in the audit exercise did not favour or impair learning.}

\textcolor{black}{In this section, we thoroughly describe the audit exercise (Section \ref{audit}) as well as the group allocation (Section \ref{allocation}). We present the chatbot (Section \ref{ChatGMP}) and avatar (Section \ref{vr}) used in the study. Finally, we present our hypotheses (Section \ref{hypotheses}) and the methods used to analyse the collected survey data (Section \ref{statistical analysis}).}

\subsection{Audit Exercise}
\label{audit}
\textcolor{black}{During the COVID-19 pandemic, the "Good Manufacturing Practice (GMP) and quality in pharmaceutical, biotech and food industry - Theoretical version" (28855) Master’s Degree course taught at the Technical University of Denmark (DTU), like many other courses around the world, was transitioned to an online format. To make the remote teaching engaging, the teachers of the course established a learning activity, which was developed as a role-playing exercise. In this activity, students were split into groups of 4-5 people and they had to simulate an audit with a fictitious pharmaceutical company, called Pharma A/S. Before the audit, the groups had to prepare a list of questions to ask during the audit exercise. The actual exercise lasted 45 minutes, where students could ask the questions they previously prepared, as well as any follow up question. Teachers usually answered by providing the documents and, when these were not available, argue why they could not be shown to the students (therefore, some answers given by the teacher might be misleading, trying to cover the non-conformities and deviations of Pharma A/S). Students' goal was to gather as many documents and information as possible in the 45 minutes given for the exercise. After the audit exercise, students had to write a report of their findings, evaluating whether the fictitious company was a valuable business partner, or some adjustments had to be made to bring the company up to standard. The learning objectives of the exercise included: the successful conduction of an audit, being able to understand GMP standard procedures and apply them to a specific use case, be critical about the provided answers and documentation, and be able to discuss the findings by writing a report.}

\textcolor{black}{In 2022, we started working on the development of an LLM that could assist the teachers in performing this exercise. In order to achieve this, we had to implement an LLM that was able to perform the same task as teachers): take part in a 45 minutes audit about the fictitious Pharma A/S company, answer students' questions about this case study and show documents supporting GMP practices adopted by Pharma A/S. We started this project by recording the audits performed by three teachers in 2022 and in 2023. This resulted in the collection of approximately 35 audits. The data was cleaned and pre-processed and transformed in a dataset of question-answer pairs. This dataset was used to enrich the knowledge of the LLM with retrieval-augmented generations (RAG) \citep{lewis2020retrieval}. In the implementation of ChatGMP between 2022 and 2024, the prompt was composed of a question (asked by the student group performing the audit) and enriched with the relevant context, containing the historical answer to the question. The context was retrieved through semantic search by calculating the cosine similarity between the input question and the most similar question in the dataset. The LLM component was based on a pre-trained, open-source model, FLAN-T5 base \citep{chung2024scaling}. The technical implementation of the original chatbot, including further details on the data collection, pre-processing and deployment, can be found in \cite{caccavale2025chatgmp} and \cite{caccavale2026large}.
}


\subsection{Student enrollment and group allocation}
\label{allocation}
An important aspect of this initiative was related to student enrollment. Before introducing ChatGMP to the course, student enrollment was limited to 120 students per semester. However, introducing \textcolor{black}{the LLM allowed us to increase student uptake without compromising on the learning activities. This was achieved through 2 years of extensive testing, and assessing that the LLM was not impeding learning. In 2026, 270 students enrolled in the course.}

\textcolor{black}{It is important to note that our intention has never been to completely replace the teachers in this exercise, only to support them and provide a tool that would allow all students to take part in this activity.} 
We also acknowledge that different individuals might have personal preferences. Since our main goal is to enable students to attend the course and, at the same time, not compromise the learning, the choice of the auditee (teacher, \textcolor{black}{chatbot} or avatar) was up to the students. \textcolor{black}{Students were asked to form groups of 4-5 people. This resulted in a total of 55 groups.} The audits had to happen within 4 weeks (3 hours per week per teacher). The number of teachers performing the audit exercise was 5. Of those, we had 3 teachers available to perform the audits physically (actively responding to students questions and showing suitable documents) and two that wanted to facilitate the AI audits. \textcolor{black}{These two teachers were present during the audit exercise to set up the audit and provide instructions.  One of the teachers facilitated the aidits with the chatbot, the other with the avatar.} We calculated the maximum amount of audits available for the teachers (3x3x4) and the ones for AI (2x3x4). For statistical purposes, we hoped to have a 50/50 split between teachers and AI (\textcolor{black}{chatbot} + avatar) audits, so we created a sign up form with the available time slots. \textcolor{black}{Since this is a group exercise, students had to agree within their group their preferred auditee. It is important to note that this decision was taken as a majority vote within each group, therefore there might have been some students who did not agree with the decision and would have preferred a different auditee.} 
We acknowledge that to rigorously validate the AI tools (both chat-based and avatar-based conversational AI models), students should have been randomly assigned to the three groups. However, we decided to prioritize the preferences and experience of the students, and their well-being and beliefs, rather than the tool validation. To strengthen and de-bias the results, we evaluate student general perspectives on AI (as shown in Figure \ref{FIG:A}).

The final distribution of audits is the following: 27 audits with a teacher, 12 audits with \textcolor{black}{the chatbot} and 16 with the avatar. Originally, we planned 3 more audits with \textcolor{black}{the chatbot}, but the LLM used within the application, Mistral (mistral-small-latest), had issues during one of the scheduled days (9 April 2026); therefore, the students performed those using the avatar. 
We decided to use a Mistral model because it is developed in France and our university is in Europe. However, after this incident, we incorporated another model, Gemini (gemini-2.5-flash), for backup, making the chatbot model agnostic and to prevent similar incidents in the future.

\subsection{Chatbot used: ChatGMP}
\label{ChatGMP}
\textcolor{black}{The chatbot designed for this course}, ChatGMP, has been used in the course since the spring semester of 2024, continuously improving the model through student feedback. 
\textcolor{black}{Compared with the previous iteration published in \cite{caccavale2025chatgmp}, the 2025 version incorporated several improvements informed by student feedback, including a redesigned interface, speech-to-text and text-to-speech functionality, and more flexible interaction. The evaluation also introduced a crossover comparison between the AI- and teacher-led versions of the audit exercise, providing a more systematic assessment of students' perceptions and experiences  \citep{caccavale2026large}.
Overall, student feedback on the AI-assisted audit exercise was positive, with students describing the chatbot as an innovative and useful tool that enabled them to work independently and receive immediate responses. However, they also identified several limitations, including the need to formulate questions precisely, occasional repetitive or insufficiently contextualised answers, and the lack of interpersonal interaction compared with a teacher-led audit \citep{caccavale2026large}.}

In this current work we focus on improving both the technical capabilities and the educational value of the system. We observed that in 2024 students were still less accustomed to interacting with an LLM, and therefore a smaller model as FLAN-T5 achieved a good performance. However, in 2025 students stated that they were using AI chatbots more consistently, and therefore we understood the need of upgrading the underlying language model, expanding its domain-specific knowledge, and enhancing its ability to handle unexpected questions and more natural conversations. The project was also extended to explore more immersive forms of interaction, including avatar-based and virtual reality environments.

\begin{figure*}
	\centering
		\includegraphics[width=\textwidth]{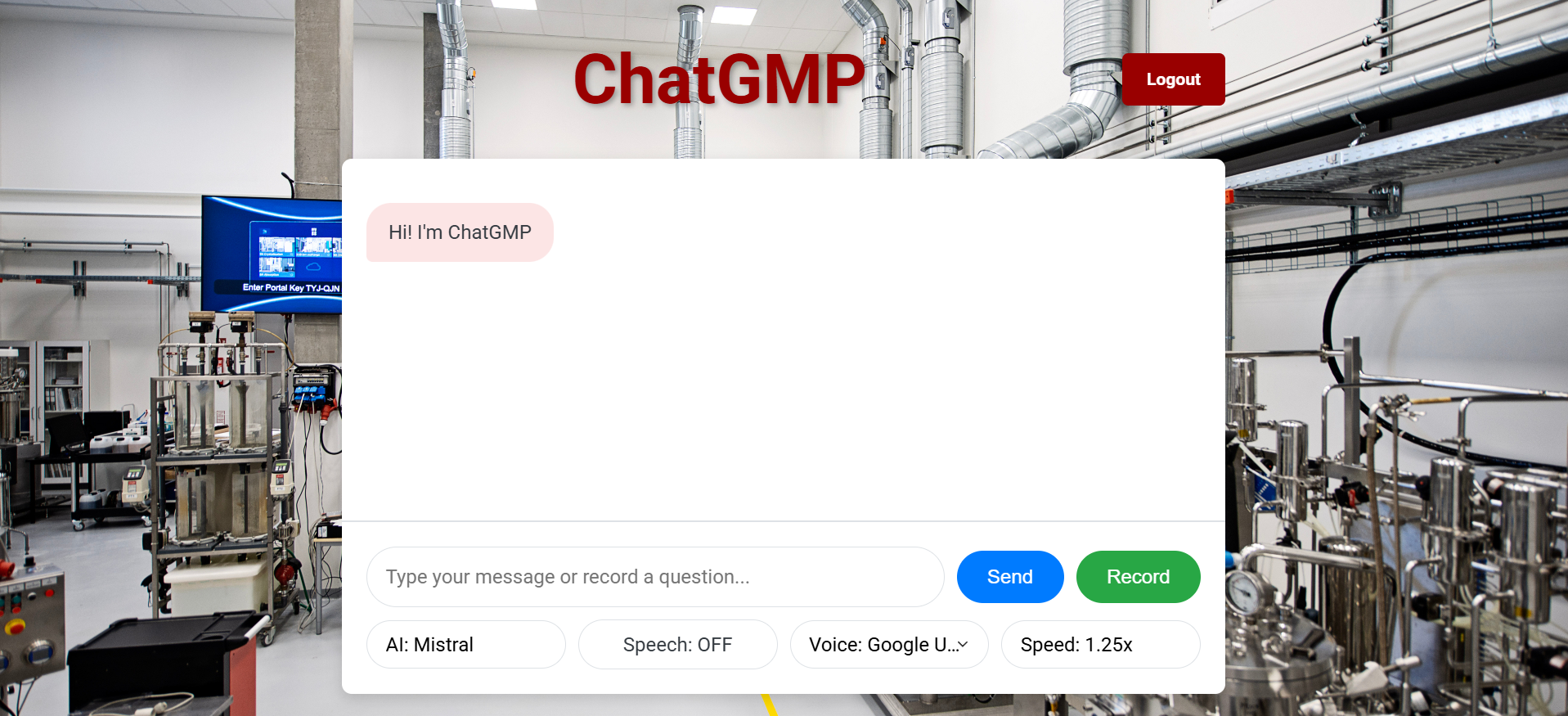}
	\caption{Front-end of the chatbot used in the course, \textit{ChatGMP}. The platform is model agnostic, students have currently two options and can adjust the options on the text read-out function.}
	\label{FIG:chatgmp}
\end{figure*}

The newest version of the ChatGMP front-end is shown in Figure \ref{FIG:chatgmp}. The application is fully deployed at \url{www.chatgmp.dk}, but login information is necessary to access. This is enabled to prevent students to repeat the exercise more than once, and therefore invalidate the results and the aim of the activity. There is an integrated speech-to-text model, Whisper \citep{whisper}, allowing students to both type their questions and speak to the chatbot. The chatbot is model agnostic, students can switch between Mistral (mistral-small-latest) \citep{mistral_models} and Gemini (gemini-2.5-flash) \citep{gemini}. They can turn on/off the text read-out, change the voice of the \textcolor{black}{chatbot}, or increase/decrease the read-out speed. In order not to overload the speech-to-text model, recordings are limited to 60 seconds per message.

\subsection{Conversational AI Avatar}
\label{vr}

After reviewing student and teacher feedback on the use of AI in parallel with teachers to enable more students to enroll in the course, we decided to make their experience more interactive. Although the chatbot version (ChatGMP) was functional, \textcolor{black}{we believe that having the possibility to interact with a human-like figure in this activity could be beneficial, resembling more the act of having a conversation with a person rather than talking to a non-embodied chatbot}. We aimed to shorten the distance between the two experiences (auditing a teacher or chatbot) by adding a "middle ground". Therefore, in the spring semester of 2026, we introduced a third option for the audit exercise, a conversational AI avatar. 

\begin{figure*}
	\centering
		\includegraphics[width=\textwidth]{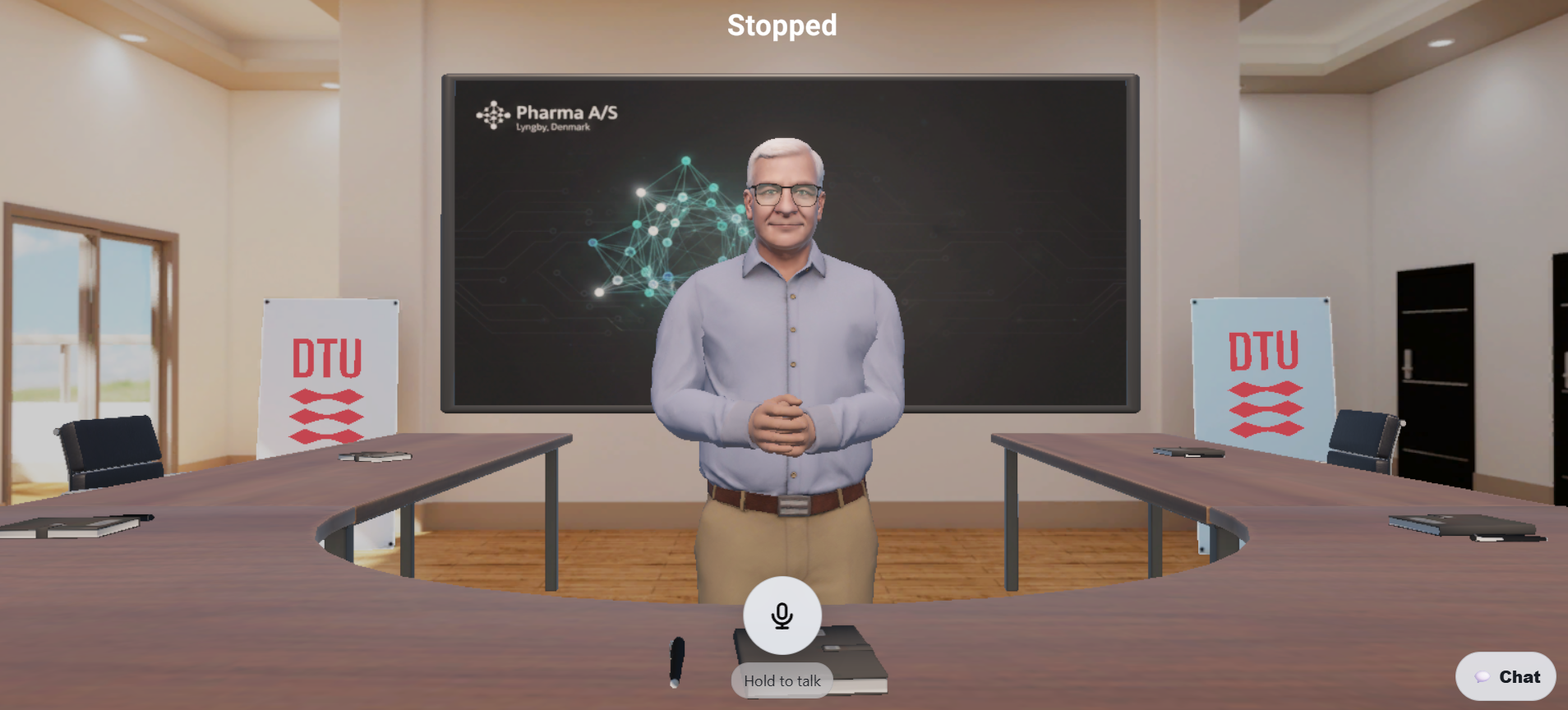}
	\caption{Conversational AI avatar use in the course (\textit{avatar}). The avatar shows when it is in standby (\textit{Stopped}) and when it is processing a question (\textit{Thinking}). On the right, students can open the chat to keep track of the conversation and the documents provided by the avatar.}
	\label{FIG:avatar}
\end{figure*}

To develop this model, we worked in collaboration with a VR company, Khora, who developed the application, including the avatar and virtual environment. 
The application is composed of multiple layers. It includes the back-end logic that uses an LLM to generate responses and retrieve the correct documents to show. The LLM used is developed by OpenAI \citep{openai} and runs on their GPU infrastructure. 
The visual layer is developed in Unity and includes the avatar/persona presented in Figure \ref{FIG:avatar} and the interactive environment surrounding it.

The experience was created in two different formats, a desktop version and an immersive VR version. In the latter, users wear goggles and are fully immersed in the experience, are able to walk around the room, and interact with multiple objects, such as the screen. The desktop version is more static, the users can speak to the avatar, open and close the presented documents but they cannot move around the room. The avatar attempts to replicate normal human behavior during conversations, such as bending slightly closer to the speaker to improve hearing.

In the course, we opted to use the desktop version, since the students were divided into groups of approximately four people, and given the limited time, it was more practical. In the bottom right, users can open the chat to keep track of the progress of the conversation or open documents that were previously shown again. 

\subsection{Hypotheses}
\label{hypotheses}
\textcolor{black}{One of the hypotheses behind the study is that conversational AI assistants could be valuable tools in this activity, supporting learning and providing an engaging environment for students.}
Additionally, we hypothesise that the avatar (or VR) version would add an improvement in the student experience from the chatbot version (the interface adds similar experience as with person). Given that (possibly) more and more students might want to register, we want to investigate if this is the right direction to improve their experience in the audit since we do not have enough physical teachers to perform the audits with all groups. Given the interactive nature of the avatar, we hypothesized students would like the experience more and perceive it as more real compared to the chatbot. 

\subsection{Survey and statistical analyses}
\label{statistical analysis}

In the survey, most items are rated on a 1–5 Likert scale. For comparisons across all three auditee cohorts, we apply the Kruskal–Wallis test. When this test indicates significance (p < 0.05), we conduct Dunn’s post hoc analysis with a Bonferroni correction. \textcolor{black}{Effect size was reported as epsilon-squared, $\epsilon_H^2 = (H-k+1)/(n-k)$, where $H$ is the Kruskal--Wallis statistic, $k$ is the number of groups, and $n$ is the total sample size.}
For pairwise comparisons (for instance, between teachers and avatar), we examine statistical differences using the Mann–Whitney U test \citep{mann1947test}. This non-parametric test is the counterpart of the t-test and is suitable for ordinal data and small sample sizes. 

For categorical questions (e.g., "Is this the future of the audit exercise?"), we employ a Chi-square test \citep{mchugh2013chi} to determine whether there is a significant association between two categorical variables. \textcolor{black}{Because expected cell counts were sparse, $p$-values were estimated using 100,000 Monte Carlo permutations while preserving group sizes and overall response totals. Effect size was reported as bias-corrected Cramér's $V$. For both effect-size measures, 95\% confidence intervals were estimated using 10,000 stratified bootstrap resamples, resampling participants within each group.}

Students could choose voluntarily whether to complete the survey; accordingly, we obtained responses from: (i) 25 students conducting the audit with a teacher, 13 with \textcolor{black}{the chatbot}, and 14 with \textcolor{black}{the Avatar-based Conversational AI}.
Because participation in the survey was voluntary, each group includes only a small number of students. As a result, true differences may not yield statistically significant results.

\textcolor{black}{Students' learning resulting from this activity is assessed in the course through a written report graded on a pass/non-pass basis. The report is submitted as a group assignment, with students working in the same groups in which they conducted the audit. Assessing learning outcomes from the written report is challenging because of its binary nature and high success rate, which provides limited detail about differences in students' achievement. In the next section, we therefore also report students' final course grades as broader academic context. Although these grades are not directly attributable to the activity, the results provide no indication that students who used the AI assistant were disadvantaged in their overall course performance.}

\section{Results}
\label{results}

\textcolor{black}{In this Section, we report the results of the study, including the students' attitudes towards AI, their perspective on the auditee they choose for audit exercise and their aggregated grades in the final exam.}

\begin{figure*}
	\centering
		\includegraphics[width=\textwidth]{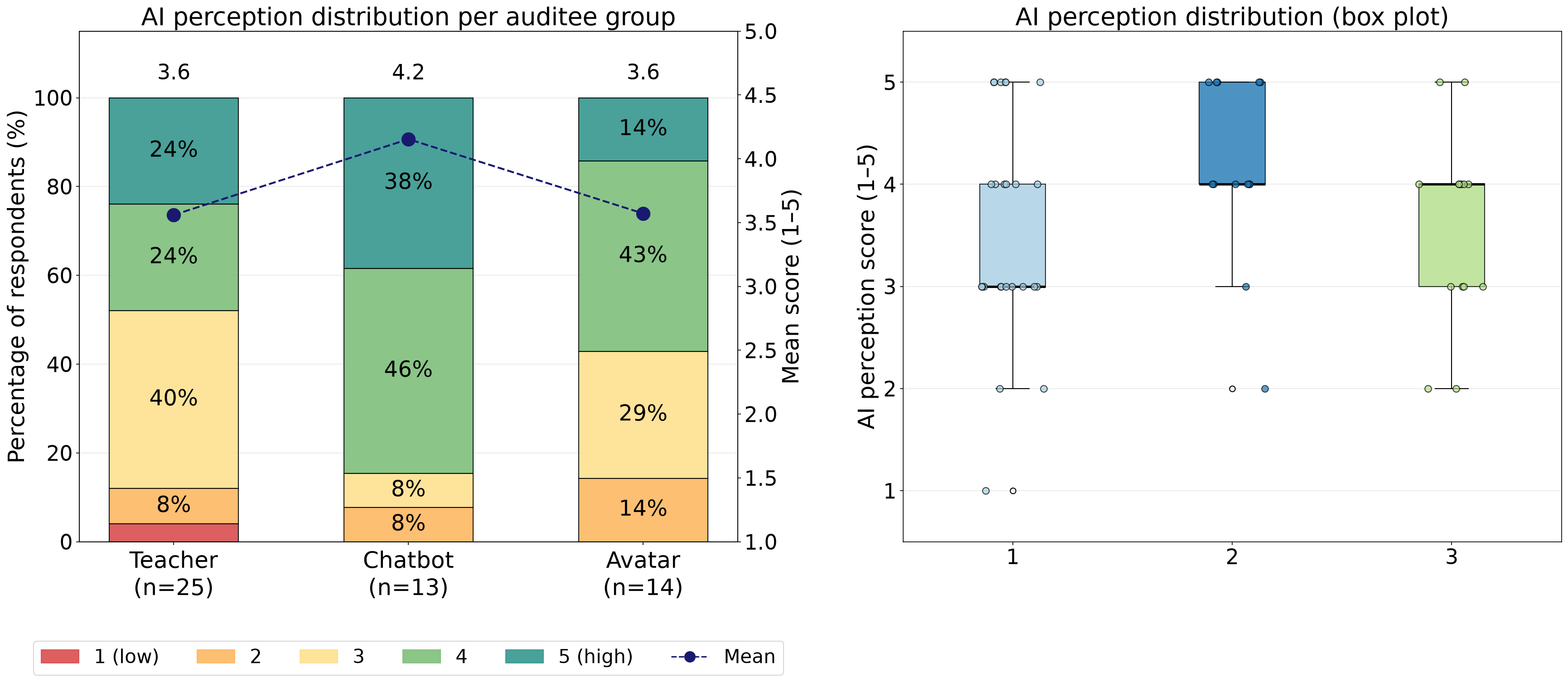}
	\caption{AI perception distribution in the different auditee groups. The left figure presents the percentages of votes in the 1-5 Likert scale and the average per group. The right figure shows the box plot distribution.}
	\label{FIG:AI_perception}
\end{figure*}

\subsection{Attitudes towards AI}
Before assessing the results of the exercise and the students perception of their auditee (teacher, \textcolor{black}{chatbot} or avatar), we assessed students attitudes towards AI, in the attempt to de-bias and provide further elements for the following analyses. 
The participants are divided into three separate groups (Teacher, Chatbot and Avatar). The analysis examines four main types of metrics: four perception-related measures rated on a 5-point Likert scale, the overall frequency of AI use, and the specific use of AI for drafting audit questions.

\begin{figure*}
	\centering
		\includegraphics[width=\textwidth]{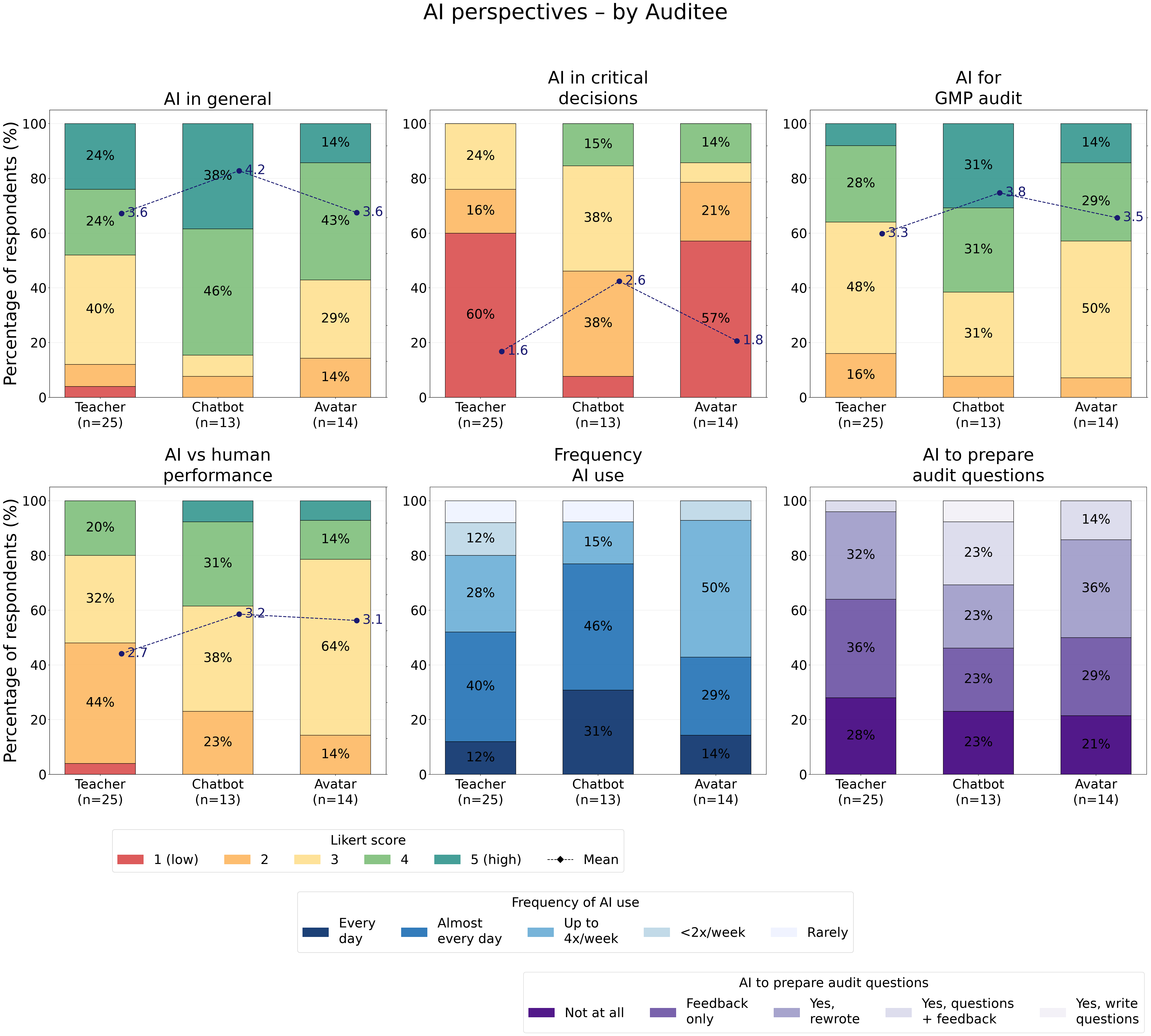}
	\caption{Distribution of the general perspectives towards AI expressed by the three cohorts (students performing the audit with the teacher, students performing the audit with \textcolor{black}{the chatbot} and students performing the audit with the avatar) for RQ1-RQ6.}
	\label{FIG:A}
\end{figure*}

Figure \ref{FIG:A} shows \textcolor{black}{how students in the three groups perceive AI. This perception is investigated through six different questions,} reported below. 

\begin{itemize}
    \item "What do you think about AI in general?" (RQ1)
    \item "How comfortable are you with AI making critical decisions (e.g., loan approval, medical diagnosis, interviewing, assessment) without human oversight?" (RQ2)
    \item "To what extent do you think AI can perform a task like the audit in the GMP course?" (RQ3)
    \item "How do you think an AI performs in this type of tasks compared to a human?" (RQ4)
    \item "How often do you use AI?" (RQ5)
    \item "Did you use AI to prepare your audit questions?" (RQ6)
\end{itemize}

For RQ1, as shown in Figure \ref{FIG:AI_perception}, as well as the fist plot in Figure \ref{FIG:A}, the higher average perception of AI (and the general trust in it) by students in the \textcolor{black}{chatbot} group is predictable compared to those who chose to audit a teacher. However, interestingly, the general regard towards AI appears to be similar between the teacher and avatar group, although the avatar uses AI. We hypothesized that the inclination towards AI would be more similar to the \textcolor{black}{chatbot} group. The Teacher group demonstrated a moderate baseline ($M = 3.6$), with 48\% of the respondents scoring 4 or 5. The Chatbot group exhibited the strongest positive sentiment ($M = 4.2$), driven by 38\% of the participants who assigned the maximum score of 5. The Avatar group decreased slightly ($M = 3.6$), although it retained a strong consensus around moderate-to-high scores (43\% at score 4). 

\textcolor{black}{For RQ2, participants in all three groups expressed reservations about AI making critical decisions without human oversight. Mean scores were lowest in the Teacher group ($M=1.6$) and Avatar group ($M = 1.8$). In both groups, most participants selected the lowest response (1): 60\% in the Teacher group and 57\% in the Avatar group. Responses in the Chatbot group were less negative ($M=2.6$); few participants selected 1, while 53\% selected 3 or 4.}

\textcolor{black}{Responses to RQ3, which asked whether AI could perform a task like the GMP course audit, were more neutral. The Teacher group had a mean score of 3.3, with 48\% selecting 3. The Chatbot group had the highest mean score ($M = 3.8$), and 62\% selected 4 or 5. The Avatar group's mean score was 3.5, with half of its participants selecting the neutral response.}

\textcolor{black}{For RQ4, unsurprisingly, participants rated AI's performance compared to a human somewhat higher in the Chatbot and Avatar groups than in the Teacher group. Mean scores were 2.7, 3.2, and 3.1, respectively. In the Teacher group, 44\% selected 2. In the Chatbot group, responses were concentrated around 3 (38\%) and 4 (31\%). In the Avatar group, 64\% selected the neutral response.}

\textcolor{black}{AI-use frequency (RQ5) also differed across groups. In the Teacher group, 12\% reported using AI every day and 40\% almost every day. Frequent use was most common in the Chatbot group, where 31\% reported using AI every day and 46\% almost every day. In the Avatar group, 14\% reported daily use and 29\% almost-daily use, while 50\% reported using AI up to four times per week.}

\textcolor{black}{Responses to RQ6, concerning the use of AI to prepare audit questions, suggest differences in how participants incorporated AI into their workflow. In the Teacher group, 28\% did not use AI for this task. The most common approaches were using it for feedback only (36\%) and rewriting its suggestions (32\%). In the Chatbot group, 23\% did not use AI, while 23\% selected each of the following approaches: feedback only, rewriting AI-generated material, and generating questions with feedback. In the Avatar group, 21\% did not use AI, 36\% reported rewriting AI-generated material, and 14\% selected the option of having AI generate the questions. This was the highest proportion for that response option among the three groups.}

\textcolor{black}{Overall, the Chatbot group reported the most frequent AI use and the highest mean scores on the perception measures reported in RQ2--RQ4. The Teacher group expressed more reservations about AI and more often reported revising AI-generated material rather than relying on it directly. The Avatar group reported frequent, though generally less-than-daily, AI use and some integration of AI into question preparation, including AI-generated drafting.}

\textcolor{black}{To be noted is that this comparison is relevant to the study's aim of examining whether AI can approximate human performance. However, because participants selected their auditee, the differences between groups should not be interpreted as effects of the audit condition; pre-existing attitudes may also have influenced that choice. As noted above, responses were collected after the audit, so participants' reported perceptions may also have been shaped by their experience. Because the survey was administered after the audit, responses about AI's ability to perform a GMP-like task may also have been influenced by participants' experience in the activity.}


\subsection{Students perception of their auditee}
\begin{figure*}
	\centering
		\includegraphics[width=\textwidth]{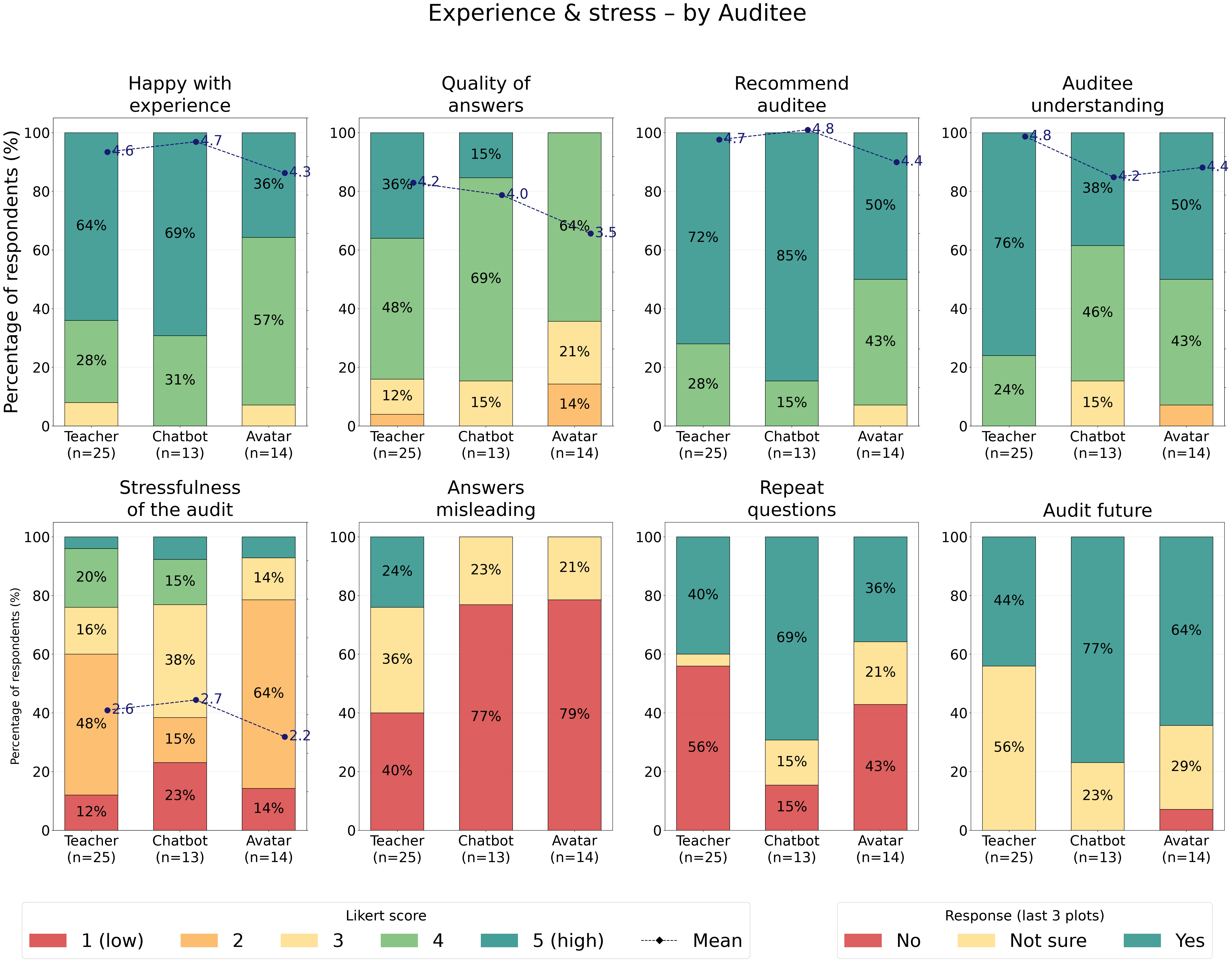}
	\caption{Distribution of the audit experience expressed by the three cohorts (students performing the audit with the teacher, students performing the audit with \textcolor{black}{the chatbot} and students performing the audit with the avatar) for RQ7-RQ14.}
	\label{FIG:B}
\end{figure*}

To systematically analyze the perceptions of the three independent cohorts (students that performed the audit with a teacher, with \textcolor{black}{the chatbot} and with the avatar), we assessed their responses in eight distinct experimental metrics. The first five subplots of Figure \ref{FIG:B} (RQ7-RQ11) analyze continuous evaluation trends via a 5-point Likert scale ranging from 1 (low) to 5 (high), while the remaining three subplots (RQ12-RQ14) evaluate nominal categorical distributions based on positive, negative or uncertain responses.

\begin{itemize}
    \item "How happy are you with your experience?" (RQ7)
    \item "After having performed the audit, what do you think of the quality of the answers?" (RQ8)
    \item "Would you recommend doing the audit with your auditee (teacher, \textcolor{black}{chatbot} or avatar) to other students?" (RQ9)
    \item "To what extent do you think that the auditee (teacher, \textcolor{black}{chatbot}, avatar) was able to understand your questions and reply coherently?" (RQ10)
    \item "How stressful was the audit?" (RQ11)
    \item "Were the answers misleading?" (RQ12)
    \item "Did you have to repeat the questions more than once to get the information that you wanted?" (RQ13)
    \item "Do you think this (e.g. audit with a teacher, \textcolor{black}{chatbot}, or avatar, according to what you choose as auditee) is the future of the audit exercise in this course?" (RQ14)
\end{itemize}

\textcolor{black}{Descriptive statistics for the ordinal questions (RQ7--RQ11) showed generally favorable evaluations of the audit experience, although response patterns varied across auditee groups (Figure~\ref{FIG:B}).}

\textcolor{black}{For RQ7, participants in all three groups reported high levels of satisfaction. The mean score was 4.6 in the Teacher group, 4.7 in the \textcolor{black}{Chatbot} group, and 4.3 in the Avatar group. In the Teacher group, 64\% selected the highest score (5) and 28\% selected 4. Corresponding proportions were 69\% and 31\% in the \textcolor{black}{Chatbot} group, and 36\% and 57\% in the Avatar group. The statistical tests did not indicate a statistically significant difference between groups ($p=0.166$), suggesting that all groups were similarly happy about their experience.}

\textcolor{black}{For RQ8, mean ratings of answer quality decreased across the groups: 4.2 for Teacher, 4.0 for \textcolor{black}{Chatbot}, and 3.5 for Avatar. In the Teacher group, 36\% selected 5 and 48\% selected 4. In the \textcolor{black}{Chatbot} group, 69\% selected 4. In the Avatar group, 64\% selected 4, 21\% selected 3, and 14\% selected 2. The omnibus test indicated a difference across groups ($p=0.035$); post hoc comparisons identified a difference between Teacher and Avatar ($p=0.029$), meaning that the quality of the answers was significantly better if auditing a teacher than the avatar.}

\textcolor{black}{RQ9 (whether the groups would recommend their auditee) received the highest and most consistently favorable ratings. Mean scores were 4.7 for Teacher, 4.8 for \textcolor{black}{Chatbot}, and 4.4 for Avatar. The proportion selecting the highest score was 72\%, 85\%, and 50\%, respectively; a further 28\%, 15\%, and 43\% selected 4. The statistical tests did not indicate a statistically significant group difference ($p=0.119$). This result is quite notable, since it suggests that the groups had a positive experience with their auditee and they would recommend it to other students. It is also interesting that the most certain group is the Chatbot group (85\% of students absolutely recommending their auditee), compared to the other two, with the avatar being the least recommended.}

\textcolor{black}{For RQ10, mean ratings of auditee understanding were high in all groups, with a lower mean in the \textcolor{black}{Chatbot} group: 4.8 for Teacher, 4.2 for \textcolor{black}{Chatbot}, and 4.4 for Avatar. The highest score was selected by 76\% of the Teacher group, 38\% of the \textcolor{black}{Chatbot} group, and 50\% of the Avatar group. The omnibus test indicated a difference across groups ($p=0.039$). There is a slight difference between the ability of the two AI systems of understanding students questions, with the chatbot being slightly worse. In fact, transcript review suggested occasional recognition errors of the speech-to-text model used in the chatbot. We do not the technical details of the Avatar's speech-recognition system; therefore, we cannot provide an in-dept system-specific explanation, only an hypothesis behind the performance difference.}

\textcolor{black}{RQ11, which assessed stressfulness, showed a different pattern. Because lower scores indicate less stress, the mean scores of 2.6 for Teacher, 2.7 for \textcolor{black}{Chatbot}, and 2.2 for Avatar reflect generally low-to-moderate reported stress. In the Teacher group, 48\% selected 2 and 12\% selected 1. In the \textcolor{black}{Chatbot} group, 38\% selected the neutral score of 3. In the Avatar group, 64\% selected 2. The omnibus test did not indicate a statistically significant difference ($p=0.425$). Although this pattern differs from the expectation that the teacher-led audit would be more stressful, familiarity with the teachers or the initial learning curve associated with the \textcolor{black}{chatbot} may have influenced students' experiences. Some students also suggested providing a live demonstration of the \textcolor{black}{chatbot} before the audit. These explanations are tentative and cannot be established from the survey data alone.}

\textcolor{black}{RQ12--RQ14 represent categorical evaluation questions.}
\textcolor{black}{For RQ12, most participants reported that the answers were not misleading. This response was selected by 40\% of the Teacher group, 77\% of the \textcolor{black}{Chatbot} group, and 79\% of the Avatar group. The remaining Teacher responses were 36\% ``Not sure'' and 24\% ``Yes''; no participants in the two AI groups selected ``Yes''. The response distributions differed across groups in the permutation test ($p_{\mathrm{perm}}=0.029$). As explained in Section \ref{audit}, the role of the teacher is also to provide non-exhaustive or slightly misleading answers, in case a document is wrongly missing, for example. Therefore, it is not negative that students find the teachers' answers to be misleading sometimes. On the other hand, it seems that the AI assistants are less likely to provide a misleading answer, since they are explicitly instructed not to hallucinate or invent facts.}

\textcolor{black}{For RQ13, 56\% of the Teacher group reported that they did not need to repeat questions, compared with 15\% of the \textcolor{black}{Chatbot} group and 43\% of the Avatar group. Conversely, 69\% of the \textcolor{black}{Chatbot} group reported needing to repeat questions, compared with 40\% of the Teacher group and 36\% of the Avatar group. The overall difference was not statistically significant ($p_{\mathrm{perm}}=0.093$). Repetition may reflect several factors, including the interaction design and speech-recognition performance, as discussed regarding RQ10.} 

\textcolor{black}{For RQ14, no participants in the Teacher or \textcolor{black}{Chatbot} groups selected ``No'', compared with 7\% in the Avatar group. Agreement that this type of audit represents the future of the course was reported by 44\% of the Teacher group, 77\% of the \textcolor{black}{Chatbot} group, and 64\% of the Avatar group. The remaining responses were ``Not sure''. The group difference was not statistically significant in the permutation test ($p_{\mathrm{perm}}=0.077$). This research question is, arguably, one of the most important in the study. These results suggest that more students performing the audit with an AI assistant believe that AI could be the future of the audit. This result could support our hypothesis that AI could be a valuable tool in this activity, aiding learning and providing an engaging environment for students.}

\begin{figure}
    \centering
    \includegraphics[width=1\linewidth]{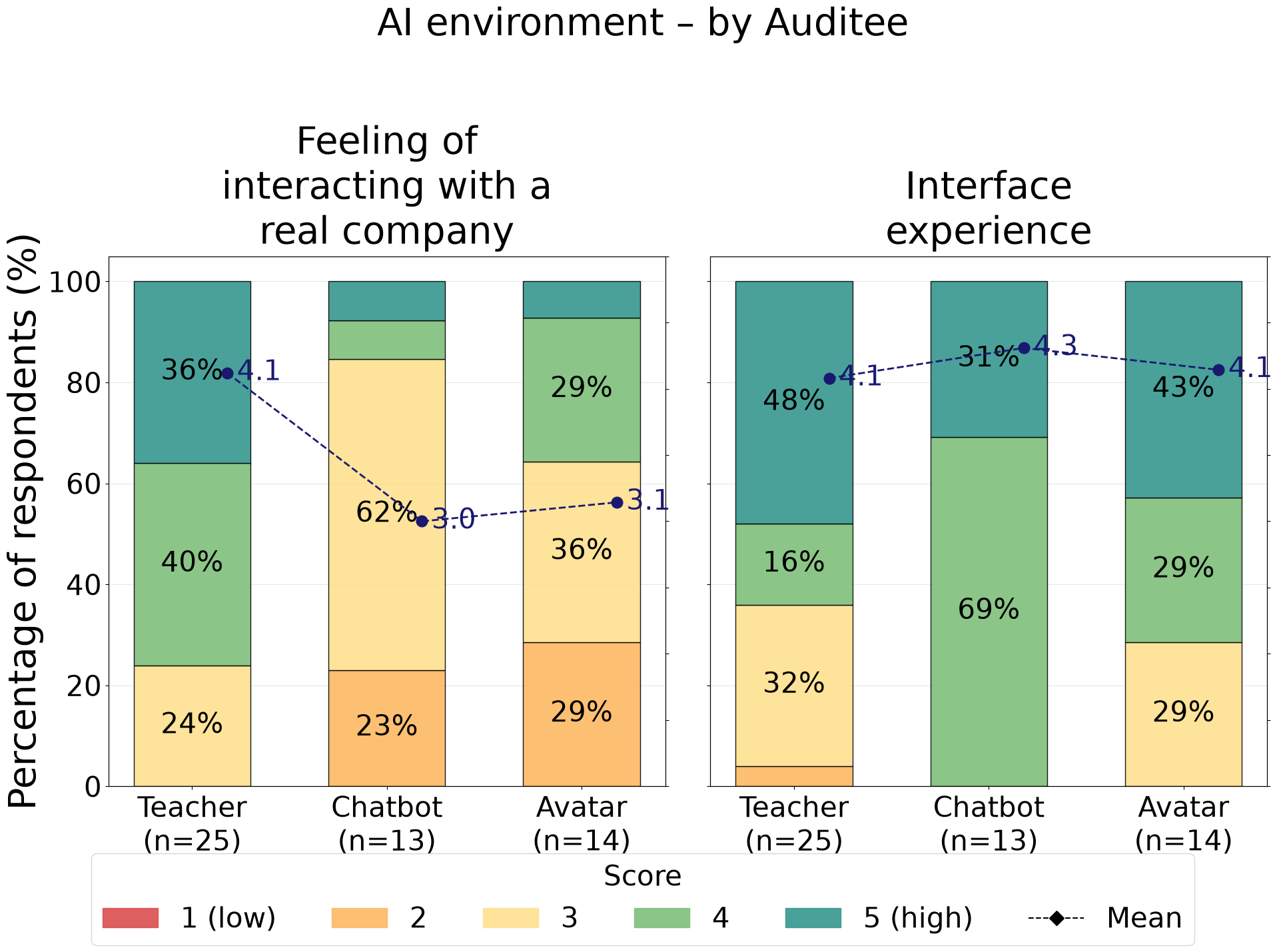}
    \caption{Distribution of the general perspectives towards AI expressed by the three cohorts (students performing the audit with the teacher, students performing the audit with \textcolor{black}{the chatbot} and students performing the audit with the avatar) for RQ15-RQ16.}
    \label{fig:company}
\end{figure}

Finally, we asked students to answer two more questions, shown in Figure \ref{fig:company}:
\begin{itemize}
    \item "To what extent did you feel like you were interacting with a real company?" (RQ15)
    \item "How would you rate interacting through the interface (if audit with \textcolor{black}{chatbot} or avatar) or the way the teachers showed the documents?" (RQ16)
\end{itemize}

Not surprisingly, students who audited a teacher felt like interacting with a real company to a higher degree compared to the others. However, the interface used seemed to be well like by all, disregarding of the mode used.

\subsection{Statistical analysis of survey responses}
Table \ref{tab:continuous_metrics} and \ref{tab:categorical_metrics} presents the statistical analysis performed on the survey responses. We can observe statistical significance in 5 questions: RQ2, RQ8, RQ10, RQ12 and RQ15.

The disposition of students performing the audit with a chatbot towards AI in critical decision (RQ2) is statistically higher than those auditing a teacher. \textcolor{black}{This result is perhaps not surprising, since we could infer that choosing to perform the audit with a text-based AI assistant instead of a teacher is perhaps an indication of a generally higher trust in AI.}

Students seem to have rated the quality of the answers (RQ8) provided by the avatar model as statistically lower than those given by the teachers. Since this was the first time this model was used in a real educational context, perhaps the back-story given to the model needs some adjustments. For example, more information regarding the nature of the exercise and the responses to be given should be incorporated in the model. The difference \textcolor{black}{between the answers given by the avatar and the chatbot} is that the latter uses RAG, and therefore might benefit from more questions being stored in the dataset provided for semantical search. In the avatar model, the information, including the role of the LLM, is provided through an elaborate back-story. However, this might be missing some responses and details. This result will be used to incorporate this data in the future and improve the model further.

Regarding RQ10, we hypothesize that this lower performance of the chatbot in understanding students' question is due to the speech-to-text model used, Whisper, which might have struggled with different accents and noise. The avatar/VR tool was developed by a partner company and not by the researchers themselves, and it is based on a pre-existing template of AI-chatbot connected to the VR environment. Therefore, we can only infer that the speech-to-text model used is different, but it is not certain.

The results for RQ12 are interesting since the exercise is originally designed to be slightly misleading, and students should investigate and ask further questions to unveil the errors in the given responses. It seems that this aspect has not been replicated by the AI tools, which tend to give more honest and flat answers.

RQ15 shows that the teacher managed to better replicate the interaction with a real company than either AI tools, however we do manage to verify our hypothesis that students would perceive the experience as more real with the avatar compared to the
chatbot.

\begin{table*}[t]
\centering
\caption{Kruskal--Wallis comparisons and effect sizes across auditees.}
\label{tab:continuous_metrics}
\scriptsize
\setlength{\tabcolsep}{3pt}
\resizebox{\textwidth}{!}{%
\begin{tabular}{lccccccc}
\toprule
& \multicolumn{3}{c}{Auditee Mean (SD)}
& \multicolumn{2}{c}{Kruskal--Wallis}
& \multicolumn{2}{c}{Effect size} \\
\cmidrule(lr){2-4} \cmidrule(lr){5-6} \cmidrule(lr){7-8}
Metric
& Teacher ($n=25$)
& \textcolor{black}{Chatbot} ($n=13$)
& Avatar ($n=14$)
& $H$
& $p$
& $\epsilon_H^2$
& 95\% CI \\
\midrule
AI in general (RQ1) & 3.56 (1.08) & 4.15 (0.90) & 3.57 (0.94) & 3.639 & 0.162 & 0.033 & [$-0.033$, 0.246] \\
AI in critical decisions (RQ2) [a] & 1.64 (0.86) & 2.62 (0.87) & 1.79 (1.12) & 9.225 & 0.010* & 0.147 & [0.018, 0.390] \\
AI for GMP audit (RQ3) & 3.28 (0.84) & 3.85 (0.99) & 3.50 (0.85) & 3.174 & 0.205 & 0.024 & [$-0.035$, 0.253] \\
AI vs.\ human performance (RQ4) & 2.68 (0.85) & 3.23 (0.93) & 3.14 (0.77) & 4.141 & 0.126 & 0.044 & [$-0.033$, 0.280] \\
Happy with experience (RQ7) & 4.56 (0.65) & 4.69 (0.48) & 4.29 (0.61) & 3.596 & 0.166 & 0.033 & [$-0.034$, 0.258] \\
Quality of answers (RQ8) [b] & 4.16 (0.80) & 4.00 (0.58) & 3.50 (0.76) & 6.724 & 0.035* & 0.096 & [$-0.010$, 0.315] \\
Recommend auditee (RQ9) & 4.72 (0.46) & 4.85 (0.38) & 4.43 (0.65) & 4.264 & 0.119 & 0.046 & [$-0.032$, 0.285] \\
Auditee understanding (RQ10) [c] & 4.76 (0.44) & 4.23 (0.73) & 4.36 (0.84) & 6.485 & 0.039* & 0.092 & [$-0.018$, 0.352] \\
Stressfulness of the audit (RQ11) & 2.56 (1.08) & 2.69 (1.25) & 2.21 (0.97) & 1.713 & 0.425 & $-0.006$ & [$-0.038$, 0.196] \\
Interacting with real company (RQ15) [d] & 4.12 (0.78) & 3.00 (0.82) & 3.14 (0.95) & 15.250 & $<0.001$* & 0.270 & [0.090, 0.521] \\
Interface experience (RQ16) & 4.08 (1.00) & 4.31 (0.48) & 4.14 (0.86) & 0.199 & 0.906 & $-0.037$ & [$-0.040$, 0.107] \\
\bottomrule
\end{tabular}%
}
\vspace{1mm}

\begin{minipage}{\textwidth}
\scriptsize
\textit{Notes:} [*] Significant at $p<0.05$. Post hoc: [a] \textcolor{black}{Chatbot} differs from Teacher ($p=0.010$); [b] Teacher differs from Avatar ($p=0.029$); [c] Teacher differs from \textcolor{black}{Chatbot} ($p=0.050$); [d] Teacher differs from \textcolor{black}{Chatbot} ($p=0.002$) and Avatar ($p=0.009$). Epsilon-squared estimates are untruncated. Confidence intervals are 95\% stratified bootstrap percentile intervals based on 10,000 resamples.
\end{minipage}
\end{table*}

\begin{table*}[t]
\centering
\caption{Categorical response distributions and effect sizes across auditees.}
\label{tab:categorical_metrics}
\scriptsize
\setlength{\tabcolsep}{3pt}
\resizebox{\textwidth}{!}{%
\begin{tabular}{llccccccc}
\toprule
Metric & Distribution
& Teacher
& \textcolor{black}{Chatbot}
& Avatar
& $\chi^2$
& $p_{\mathrm{perm}}$
& $V_{\mathrm{bc}}$
& 95\% CI \\
\midrule
\textbf{Frequency AI use (RQ5)} & Every day & 12\% & 31\% & 14\% & 7.874 & 0.468 & 0.000 & [0.000, 0.437] \\
& Almost every day & 40\% & 46\% & 29\% & & & & \\
& Up to 4 times/week & 28\% & 15\% & 50\% & & & & \\
& Less than 2 times/week & 12\% & 0\% & 7\% & & & & \\
& Rarely & 8\% & 8\% & 0\% & & & & \\
\midrule
\textbf{AI to prepare questions (RQ6)} & Not at all & 28\% & 23\% & 21\% & 6.871 & 0.600 & 0.000 & [0.000, 0.421] \\
& Only for feedback & 36\% & 23\% & 29\% & & & & \\
& Re-wrote answers & 32\% & 23\% & 36\% & & & & \\
& Direct inspiration & 4\% & 23\% & 14\% & & & & \\
& Fully copy-pasted & 0\% & 8\% & 0\% & & & & \\
\midrule
\textbf{Answers misleading (RQ12)} & No & 40\% & 77\% & 79\% & 10.451 & 0.029* & 0.253 & [0.093, 0.436] \\
& Not sure & 36\% & 23\% & 21\% & & & & \\
& Yes & 24\% & 0\% & 0\% & & & & \\
\midrule
\textbf{Repeat questions (RQ13)} & No & 56\% & 15\% & 43\% & 7.957 & 0.093 & 0.197 & [0.000, 0.445] \\
& Not sure & 4\% & 15\% & 21\% & & & & \\
& Yes & 40\% & 69\% & 36\% & & & & \\
\midrule
\textbf{Audit future (RQ14)} & No & 0\% & 0\% & 7\% & 7.423 & 0.077 & 0.183 & [0.000, 0.480] \\
& I am not sure & 56\% & 23\% & 29\% & & & & \\
& Yes & 44\% & 77\% & 64\% & & & & \\
\bottomrule
\end{tabular}%
}
\vspace{1mm}

\begin{minipage}{\textwidth}
\scriptsize
\textit{Notes:} [*] Monte Carlo permutation $p<0.05$, based on 100,000 permutations preserving group sizes and overall response totals. $V_{\mathrm{bc}}$ is bias-corrected Cramér's $V$. Confidence intervals are 95\% stratified bootstrap percentile intervals based on 10,000 resamples.
\end{minipage}
\end{table*}

\subsection{Students' performance}
\begin{table}[htbp]
\centering
\caption{Your table caption.}\label{tbl1}
\caption{Aggregated results showing the pass rate \textcolor{black}{in the final exam}. The first set of percentages is calculated based on the overall number of students enrolled, the second \textcolor{black}{reports the percentage distribution} based on the specific audit choice (e.g. teacher, \textcolor{black}{AI} or no audit).}\label{tbl1}
\begin{tabular*}{0.9\linewidth}{@{\extracolsep{\fill}} llll @{}}
\toprule
\textbf{Result Total} & \textbf{Teacher \%} & \textbf{AI \%} & \textbf{N/A \%}\\
\midrule
Passed & 45.1 & 44.3 & 1.89 \\
Non-passed & 3.4 & 4.9 & 0.38 \\
\bottomrule
\textbf{Result Relative} &  &  & \\
\midrule
Passed & 93.0 & 90.0 & 83.3 \\
Non-passed & 7.0 & 10.0 & 16.7 \\
\bottomrule
\end{tabular*}
\end{table}

The learning from the audit exercise was assessed through a written report, which \textcolor{black}{all} the students passed. One of the teachers observed that in the report, students conducting the physical audits focused more on the written documents that they obtained during the audit and not on the spoken words (where the teacher had to make one group revise the report as they did not include anything said/explained during the audit). On the other hand, in the reports handed in by the \textcolor{black}{chatbot}/avatar groups, they also include the answers from the chatbot as they received them as a transcript. 

\textcolor{black}{We present the results of the final exam in Table \ref{tbl1}. To be noted is that the result of the exam is not directly related to the audit exercise, however it can provide a measure that the different auditee did not harm learning}. The final exam was multiple choice and students had to study general GMP procedures, not related to the fictional company audited in the audit exercise. However, we still consider it relevant to report and discuss the general performance of the students in the final exam with respect to their choice of auditee.

To measure the statistical significance of these results, we run a Chi-Square test. The value ($\chi^2$) found is 1.203, with 2 degrees of freedom (df) and a p-value of 0.548. Since the p-value ($0.548$) is above the standard significance threshold of $0.05$, there is no statistically significant difference among the three groups (teacher, AI and N/A). Even though the N/A group has a lower pass rate ($83.3\%$), its sample size is small (only 6 students total). Because of this small sample size, the mathematical model views the variance as purely random fluctuation rather than a systemic trend.

Even when we break the AI tools down into their specific sub-types (\textcolor{black}{chatbot} vs avatar), the overall p-value is 0.181. Because this is still above the 0.05 significance threshold, the statistical conclusion remains the same: there is no statistically significant difference in student pass rates across the Teacher, \textcolor{black}{chatbot}, or avatar groups. While \textcolor{black}{the chatbot}'s pass rate seems lower (84.9\% compared to the others at ~93\%), the sample size for that group (53 students) is small enough to conclude that this gap as likely due to random variation rather than a definitive failure of the tool.

\begin{table}[htbp]
\caption{Individual teachers differences.}\label{tbl2}
\begin{tabular*}{0.9\linewidth}{@{\extracolsep{\fill}} lllll @{}}
\toprule
\textbf{Teacher} & \textbf{AI} & \textbf{Total} & \textbf{Passed [\%]} & \textbf{Non-passed [\%]}\\
\midrule
A & No & 41 & 37 [90.2] & 4 [9.8]  \\
B & No & 48 & 44 [91.7] & 4 [8.3]  \\
C & No & 39 & 38 [97.4]  & 1 [2.6] \\
D & Yes & 55 & 50 [90.9]& 5 [9.1 ] \\
E & Yes & 75 & 67 [89.3] & 8 [10.7] \\
\bottomrule
\end{tabular*}
\end{table}

\textcolor{black}{It is also worth mentioning that the groups performing the audit with a teacher were randomly assigned to one of three different teachers. Therefore, while we assume that the type of auditor (teacher, chatbot or avatar) could have an impact on the students' experience and their performance, it is also fair to report the difference among the teachers in how many students passed or failed. This investigation is reported in Table \ref{tbl2}.}

To validate the results in Table \ref{tbl2} and measure if there is any statistical difference between the teachers (i.e. to answer the question \textit{"Is there any relationship between who the teacher is and whether a student passes or fails?"}), we performed a Chi-Square test. The p-value is higher than the threshold 0.05 (0.68), therefore, there is no statistically significant difference. Even though Teacher C has a 97.4\% pass rate and Teacher E has an 89.3\% pass rate, the sample size is too small or the variation is too tight to say they are performing differently. To further analyze the results, since the Chi-Square test is more robust with values above 5, we also performed a Fisher Exact Test between Teacher C and E. Results are still not statistically significant.

\section{Discussion}
\label{discussion}
\textcolor{black}{This section aims to contextualize the results presented in this study, presenting the key patterns and implications of the study, our perspective on automating educational processes and scalability and industrial transferability of this work.}

{\color{black}
\subsection{Key patterns and implications} 

In Section \ref{results}, we showed that there is a correlation between the chosen auditee and students' perception and trust in AI. In fact, the three groups differed in their reported perceptions of AI and in how they incorporated it into audit-question preparation. The Teacher group reported frequent general AI use, but expressed greater reservations on the perception measures and tended to use AI for feedback or to revise its suggestions. The Chatbot group reported the most frequent use and the highest mean scores across the perception measures discussed above. The Avatar group showed regular AI use and practical integration into question preparation, while its responses on the perception measures were generally moderate. The Chatbot and Avatar groups gave similar mean ratings for RQ4, which asked how AI performs on these tasks compared with a human. 

Answers to RQ7 and RQ14 are among the most notable results for this study.
RQ7 investigates the overall happiness of students regarding the experience ("How happy are you with your experience?"). Here, all groups have an above-average satisfaction, with the Chatbot group being the absolute happiest, followed by the Teacher group and the Avatar group. This is meaningful for the study because it verifies our hypothesis that using Conversational AI is a valuable tools in this activity, providing an overall positive experience for students. RQ14, ("Do you think this, e.g. audit with a teacher, chatbot, or avatar, according to what you choose as auditee, is the future of the audit exercise in this course?") is equally interesting, since it investigates whether students think AI could be the future of the audit exercise. Students in the Chatbot group are the most sure that the chatbot could effectively perform this activity in the future, followed by the Avatar group and then, surprisingly, the Teacher group. This result might be due to the fact that students are also aware that this activity would be impossible for the teachers to perform without AI given the lift of the enrollment limit, and therefore they might recognize the benefit it might have. To be noted is that our aim is not to replace the role of teachers in this activity, only to provide viable alternatives for the groups that are comfortable interacting with AI. 
Taken together, the findings suggest that conversational AI, both in text-based and avatar-based form, may be a feasible and generally acceptable auditee for this activity. Participants in all three groups reported high satisfaction and willingness to recommend their auditee. 

However, the AI tools seems to be having more problems with technical aspects such as the quality of the responses (RQ8) and understanding questions (RQ10). Differences in ratings of answer quality and understanding suggest that acceptability and perceived interaction quality are distinct dimensions. 
Differences were more apparent between auditees in evaluations of answer quality (RQ8) and auditee understanding (RQ10). 
The avatar-based conversational AI assistant received the lowest scores for answer quality, showing significantly worse answers than the ones given by the teachers. 
These findings suggest that evaluations may depend on specific response characteristics (since quality of the answers given by the chatbot is high), rather than simply on whether the auditee was human or AI.

Interestingly, the more stressful medium for students seemed to be the chatbot. This aspect should be further investigated.

The categorical responses add further nuance. Participants in the \textcolor{black}{Chatbot} and Avatar groups were more likely than those in the Teacher group to report that answers were not misleading, which is an intended aspect of the exercise. In fact, the auditee should try to convince the auditor that not having an important document is not much of a problem (while in reality students should be able to understand that it is). Another issue with the AI assistants is represented by the higher observed repetition rate in the \textcolor{black}{Chatbot} group, which may reflect interaction or speech-recognition challenges. However, as previously discussed, the survey data cannot establish the cause. 

These findings suggest practical priorities for AI supported audit activities. A short demonstration before the exercise may help students become familiar with an unfamiliar interface, particularly under time constraints. Evaluation should also distinguish overall acceptance from task-specific performance, including answer comprehensibility, speech recognition, and students' ability to identify misleading information.

Several limitations constrain interpretation. Participants selected their auditee, so pre-existing preferences or attitudes may have contributed to the observed group differences. The survey was administered after the audit, meaning responses may reflect the activity experience as well as prior perceptions of AI. The sample was modest, and many effect-size confidence intervals were wide. Categorical outcomes were also sparse, so permutation-based $p$-values were used instead of relying on the asymptotic chi-square approximation. These findings should be interpreted as preliminary findings and not as pedagogical evidence that auditee type caused differences in students' perceptions or learning. Future research will use larger samples and controlled comparisons, alongside direct measures of learning, answer accuracy, speech-recognition performance, and students' ability to identify misleading information.

Ultimately, the results suggest that while AI and virtual tools match traditional human-led benchmarks in high-level outcomes, their integration introduces distinct variables that warrant nuanced instructional design. Academically, whether the students had a human teacher, \textcolor{black}{chatbot}, or avatar, the final pass rates were statistically tied (90\% to 93\%). At a high level, the tools are equally effective. However, the way students felt during the process varied: students felt a significantly stronger sense of \textit{"interacting with a real company"} with the human teacher and had to repeat their questions more to the chatbot to get the answer needed. 
In summary, although the preliminary results seem promising, especially regarding the satisfaction of students with the AI tools (RQ7 and RQ14), adjustments to the tools, and how these are implemented in the course should be made to ensure even more alignment of the AI experience with the human-led counterpart. Since the quality of responses of the chatbot is perceived to be high by the students, and back-story provided to the avatar-based conversational AI is built on the same data, we believe that improving the flow of information injected in the back-story will lead to an overall increase in quality of responses. Therefore, while we could not verify our hypothesis that the avatar would add an improvement in the student experience from the chatbot in regards to quality of learning at this time, we are certain that we are going in the right direction to improve their experience. Therefore, by incorporating the feedback received, this can be improved and verified in the next semester of the course.
}

\subsection{Automating educational processes}
\textcolor{black}{This study aims to provide evidence that automating certain educational processes through the use of conversational AI could lead to an overall improvement in student experience without compromising learning.} 
As shown in Section \ref{results}, while the students show different stages of readiness to fully embrace AI, by allowing them to choose whether they prefer to interact and receive feedback from a teacher or AI tool, we were able to increase the enrollment of the course from 120 to 270 students without compromising the overall quality of the course. \textcolor{black}{Being able to lift the enrollment upper limit is a significant improvement for students, who can now freely enroll in the course rather than having to chose another elective or postponing it to the next year, potentially even delaying their graduation.} 

In general, we believe that role play exercises represent an effective approach to mimic real world scenarios. Using technologies and strategies such as conversational AI avatars and VR, as well as gamification, we could increase student engagement and lead to an improvement in learning. Therefore, technologies like the ones presented in this study, if implemented responsibly and used correctly, could lead to a better overall experience for students. Referring to Bloom's Taxonomy \citep{bloomtaxonomy}, similar tools should be evaluated on the \textit{"Apply"} level: evaluating how well do simulated tasks (e.g. the audit exercise) prepare students for a real task, and, as importantly, how well do students they feel prepared. 

\textcolor{black}{Finally, how a course is structured needs to be adjusted depending on the technologies used. Having an interview-like exercises talking in person with a human teachers is substantially different than typing through a computer. Therefore, since new technologies such as chatbots and avatar based-conversational AI provide different experiences, the activities and learning objectives need to be adjusted. These adjustments can involve training teachers to use the technologies, having a lecture introducing the tools to students, created guidelines on how to prompt and interact with these models.}

\subsection{Scalability and industrial transferability}
Outside traditional educational settings, AI is reshaping industrial operations through smarter process control, predictive maintenance and digitalization, while simultaneously changing job roles and required skills \citep{babashahi2024ai}. AI can be used to customize onboarding programs, tailoring content by role and using chatbots/virtual assistants for rapid Q\&A, while augmenting human interaction \citep{brown2024impact}. Broader workforce reviews emphasize comprehensive, inclusive AI upskilling and lifelong professional development as essential for readiness in AI‑intensive industries \citep{joshi2025agentic}. This work focuses on the application of LLMs in an educational context. However, similar AI-powered tools, can be easily applied to any other applications, including industrial processes. 

The course used as testing ground is in GMP. GMP represents an important field within chemical engineering, and the automation of repetitive tasks could widely benefit the industry. For example, the AI-powered chatbot could be used in industrial GMP contexts by simply changing the input data and documents to which the model has access.

To broaden the scope of this research, we posit that the methodology demonstrated herein holds significant potential for scalable application across diverse industrial sectors characterized by repetitive workflows. 
A primary candidate for this technology is the pharmaceutical industry, specifically within Quality Control (QC) operations. In this domain, personnel routinely perform highly structured tasks, such as generating standardized documentation to log procedural deviations. The integration of conversational AI into these frameworks could offer a viable pathway for automating routine reporting, thereby optimizing workforce allocation and yielding substantial reductions in operational labor costs.
The modular design of our system supports the inclusion of "human-in-the-loop" verification, where an employee can leverage the tool to speed up the review of documentation, e.g. batch records, while still retaining full and final decision-making responsibility, rather than delegating it to the AI tool.

\textcolor{black}{We also believe that avatar-based Conversational AI and VR technology can be a valuable asset to be used during training and onboarding in industrial context. 
These tools could be used, for example, to train before performing tasks in the laboratories, where the employees interact with a virtual replica of the real laboratory or pilot plant.}

Finally, the deployment of both AI and VR technologies requires a comprehensive sustainability assessment. It is imperative to evaluate whether the operational efficiencies and advancements gained through AI integration yield a net positive return when critically balanced against its resource expenditure, particularly with respect to energy consumption and carbon footprints.

\section{Conclusion}

Our findings reveal that while overall student happiness, task-specific metrics, and future recommendations remained highly robust and statistically comparable across all three cohorts, critical points of divergence emerged regarding structural execution and qualitative user experiences. Specifically, the data indicate distinct trade-offs between modalities: \textcolor{black}{the groups that performed an audit with a teacher} reported significantly higher occurrences of perceived misleading information (which is intended as part of the exercise) and an enhanced sense of authentic contextual immersion. The perceived quality of the answers given by the AI tools (\textcolor{black}{chatbot} and avatar) was slightly lower than the teacher's answers; additionally, students performing the audit with the AI tools also expressed concerns about occasional algorithmic hallucinations and incorrect answers. On the other hand, the certainty of these cohorts (performing the audit with either the chatbot or the avatar) that the use of AI tools should be the future direction of the exercise is greater \textcolor{black}{and arguably the most notable result of this study.}

\textcolor{black}{The use of AI in this activity led to a raise of the enrollment limit in the course, allowing a 66.6\% increase in student uptake yearly. Although only one of these tools, the avatar, was introduced in this work, and we cannot claim that the increase in student enrollment is a direct consequence of this study, continuously performing research on the topic and improving these tools will help keeping the student enrollment unrestricted, without compromising the teaching quality or reducing the activities planned for the course}. 

Ultimately, our findings suggest that by meticulously aligning these technologies with carefully defined learning objectives, institutions can leverage the potential of AI to enhance \textcolor{black}{learning activities}, not only in higher education, but also within the broader demands of the modern industrial landscape, ensuring that the pursuit of innovation remains firmly anchored in academic rigor and pedagogical responsibility.

\section{Future perspectives}
\label{future}
In the future, we will continue to develop on the tools created and investigate more deeply the pedagogical implications of using AI in this, and similar, contexts. Moreover, we will include all feedback received by the students to further improve their experience.

Additionally, to improve the full immersion of the experience, we will expand on the VR capability of the tool, by incorporating the possibility to show actual photos and schematics of the pilot plant that the exercises uses for inspiration (at DTU). Finally, we will also simulate a visit to the pilot plant in a VR environment, which could include some hygienic design investigations.


\section{Limitations}
\label{limitations}

Data collection is limited to students who volunteered to provide their feedback on the course and the initiative. \textcolor{black}{We acknowledge that the survey had a relatively small number of responses. This may limit how representative the findings are and could introduce non-response bias if respondents differ systematically from those who did not participate. Therefore, the results should be interpreted with caution, and future studies should aim to recruit a larger and more representative sample.} 

Although the results reported see the \textcolor{black}{avatar-based conversational AI} mode to be the less preferred by the students and the one producing the lowest quality of responses, we believe this is attributable to the application being created in early 2026 and used in March and April 2026. Therefore, not all issues with the responses were solved in the initial testing phase before using the AI assistant in the course. We are certain that this can be improved in the future by increasing the knowledge of the model in the "background story". 

We recognize that a rigorous validation of the AI tools would have required randomly assigning students to the three groups (teacher, chatbot and avatar). Nonetheless, we chose to prioritize students’ preferences, prior experiences, well-being, and beliefs over strict tool validation. To reduce bias in our findings, the results should be interpreted by taking into consideration students' overall attitudes toward AI (see Figure \ref{FIG:A}).

As already mentioned in Section \ref{audit}, the LLM used in \textcolor{black}{the chatbot} had problems (possibly due to the increased amount of traffic) during one of the exercise days. The students decided to then perform the audit with the avatar (also AI-powered). We asked students how they felt about the experience after the audit and they expressed that they did not mind the change. However, we cannot exclude the possibility that this incident might have affected the experience and perceptions of these students.

\section*{Availability of code and materials}
\label{code}
This study is committed to the open-source set of principles. The code for the chatbot, ChatGMP is freely available on GitHub at \url{https://github.com/FiammettaC/ChatGMP}. The code is published under the MIT license. Currently, it is compatible with Python 3.x, for Windows and Linux operating systems. It is also recommended to use a conda environment to run the code. 

\textcolor{black}{The technical implementation of the Avatar-based Conversational AI Assistant was developed by Khora. Although the source code and virtual environment are not publicly available, the system infrastructure is described in Section \ref{vr}. Further information may be requested from the corresponding author.}

\section*{Funding}
Funding for this study was provided by (i) the European Union under Horizon Europe research and innovation program with the grant agreement 101159993, within the HORIZON-WIDERA-2023-ACCESS-02-0 call (Dig4Bio); (ii) Novo Nordisk Foundation, the grant number is: \newline NNF22OC0080136 - Real-time sustainability analysis for Industry 4.0 (Sustain4.0); and (iii) Technical University of Denmark (DTU).

\section*{Statements on ethics and open data}
Informed consent was obtained from all participants and their privacy rights were strictly observed. Before participating, all participants were informed that the collected data would be used for research purposes and would be handled anonymously. Detailed information on the study was provided to the participants before starting the survey, reassuring them of their right to voluntary participation and withdrawal. The data can be obtained by sending an e-mail request to the corresponding author.

\section*{Declaration of competing interest}
The authors declare that they have no known competing financial interests or personal relationships that could have appeared to influence the work reported in this paper.



\section*{Declaration of generative AI and AI-assisted technologies in the manuscript preparation process}

During the preparation of this work, the author(s) used GPT-6 Luna through GitHub Copilot in order to modify the Python code used to generate the plots and to rephrase text. After using this tool/service, the author(s) reviewed and edited the content as needed and take(s) full responsibility for the content of the published article.


\bibliographystyle{cas-model2-names}

\bibliography{cas-refs}

\end{document}